\documentclass[conference]{IEEEtran}
\IEEEoverridecommandlockouts
\usepackage{cite}
\usepackage{amsmath,amssymb,amsfonts}
\usepackage{algorithmic}
\usepackage{graphicx}
\usepackage{textcomp}
\usepackage{xcolor}
 \usepackage{multirow}
 \usepackage{fancyhdr}
\def\BibTeX{{\rm B\kern-.05em{\sc i\kern-.025em b}\kern-.08em
    T\kern-.1667em\lower.7ex\hbox{E}\kern-.125emX}}
\begin{document}

\title{WDA-Net: Weakly-Supervised Domain Adaptive
Segmentation of Electron Microscopy\\
}

\author{\IEEEauthorblockN{Dafei Qiu}
\IEEEauthorblockA{\textit{Xiamen Key Laboratory of} \\\textit{Computer Vision and Pattern Recognition} \\
\textit{Huaqiao University, China}}
\and
\IEEEauthorblockN{Jiajin Yi}
\IEEEauthorblockA{\textit{Meitu, Inc.} \\
\textit{Xiamen, China}}
\and
\IEEEauthorblockN{Jialin Peng*}
\IEEEauthorblockA{\textit{College  of Computer Science and Technology} \\
\textit{Huaqiao University, China}\\
2004pjl@163.com}
\thanks{ The first two authors  contributed equally. }
\thanks{ *Correspondence author.}
}

\maketitle
%\thispagestyle{fancy}
%\cfoot{978-1-6654-6819-0/22/\$31.00~\copyright~2022 IEEE}
%\lfoot{}
%\renewcommand{\headrulewidth}{0mm}
\begin{abstract}
Accurate segmentation of  organelle instances is  essential for electron microscopy analysis. Despite the outstanding performance of fully supervised methods, they highly rely on  sufficient  per-pixel annotated data and are sensitive to  domain shift.
Aiming to develop a highly  annotation-efficient approach with competitive performance, we focus on weakly-supervised domain adaptation (WDA) with a  type of extremely sparse and weak annotation demanding minimal annotation efforts, i.e., sparse point annotations on only a small subset of object instances.  To reduce  performance degradation arising from domain shift, we explore multi-level transferable knowledge through conducting three complementary tasks, i.e., \textit{counting}, \textit{detection}, and \textit{segmentation}, constituting a \textit{task pyramid} with different levels of domain invariance. The intuition behind this is that after investigating a related source domain, it is much easier to spot similar objects in the target domain than to delineate their fine boundaries. Specifically, we enforce counting estimation as a global constraint to the detection with sparse supervision, which further guides the segmentation. A cross-position cut-and-paste augmentation is introduced to further compensate for the annotation sparsity. Extensive validations show that  our model with only 15\% point annotations can achieve comparable performance as supervised models and shows robustness to annotation selection.
\end{abstract}


\section{Introduction}
\begin{figure}[t]
\centering
\includegraphics[width=0.49\textwidth]{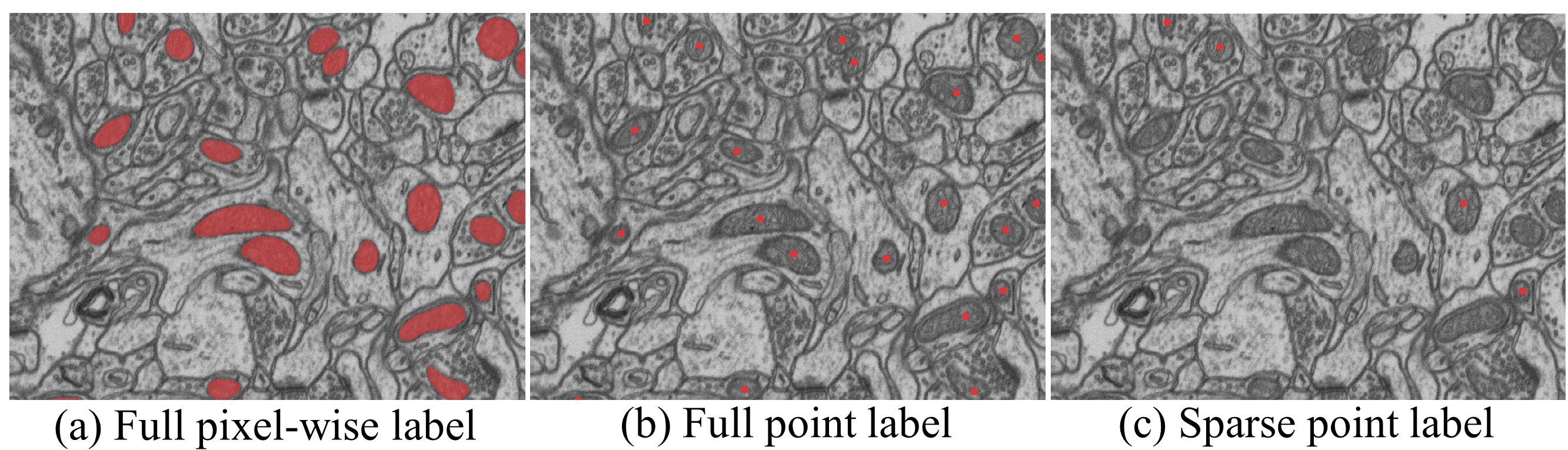}
\caption{Comparison of the introduced sparse point annotation with other types of annotations for mitochondria in EM images.
} \label{fig:1}
\end{figure}

Segmenting  subcellular organelles, such as mitochondria instances, from large-scale  electron microscopy (EM) (Fig. \ref{fig:1}) is an indispensable task for many neuroscience research and clinical studies \cite{nunnari2012mitochondria}.  While state-of-the-art automatic segmentation methods are based on  supervised learning  \cite{ronneberger2015u,peng2021medical}, the widespread label scarcity and  data distribution shift (also known as \textit{domain shift}) in practical scenarios usually  preclude their applications.  Particularly,  EM images of different tissues/species show significantly different contents and appearances, which result in large domain shift. The various types of electron microscopes further increase the image diversity.  While supervised training on each domain will reach highest performance, collecting sufficient  data labeled by experts for model training is usually prohibitive.

 To alleviate  the heavy annotation burden and label scarcity on an unexplored target domain, a promising solution is to conduct domain adaptation (DA), which leverages a related but well annotated  source domain and  bridges the domain gap by learning domain-invariant representations  \cite{DAN}. Recently,  unsupervised domain adaptation (UDA), assuming completely no labels on the target domain, has been extensively developed and also has made impressive progress \cite{DANN,Outputspace,DeepG,peng2020unsupervised} on various tasks. However, for most challenging segmentation tasks that involve high-dimensional and structured prediction,  UDA approaches are  far from practical usage due to the significantly low performance compared to  the fully supervised counterparts. One direct way to promote the performance is  to  conduct semi-supervised domain adaptation (SDA)\cite{chen2021semi}. However, this strategy will significantly increase data annotation burden  and delay the model deployment. Moreover, dense annotations may contain redundancy, especially for delineating  subcellular organelle instances. For practical usage, it is desirable to devise a domain adaptive segmentation method that can achieve sufficiently high performance with least annotation cost.

In this study,  we consider a novel class of \textit{weakly-supervised domain adaptation (WDA)} setup, assuming that the target training data has sparse center-point annotations on a randomly-sampled  small proportion (e.g., 15\%) of mitochondria instances. Compared to fully pixel-wise  annotation and full point annotation shown in  Fig. \ref{fig:1},  our  \textit{sparse point annotation} is extremely efficient and can be accomplished by non-expert with minimal cost in several minutes. However, this WDA setup is  very  challenging in comparison with fully-supervised and semi-supervised settings due to the \textit{sparse instance coverage}, \textit{missing information about object boundaries and appearances}, and \textit{the only availability of labels for the positive class}. Compared to other weakly-supervised scenarios \cite{peng2021medical} that assume the availability of location cues for all instances (e.g., bounding boxes \cite{2021Box}, scribbles \cite{dorent2020scribble}, and full center points \cite{Nishimura2021} for all instances), this sparse point annotation setup is also much more challenging due to missing location cues for most instances and missing background information.
\begin{figure}[t]
\centering
\includegraphics[width=0.48\textwidth]{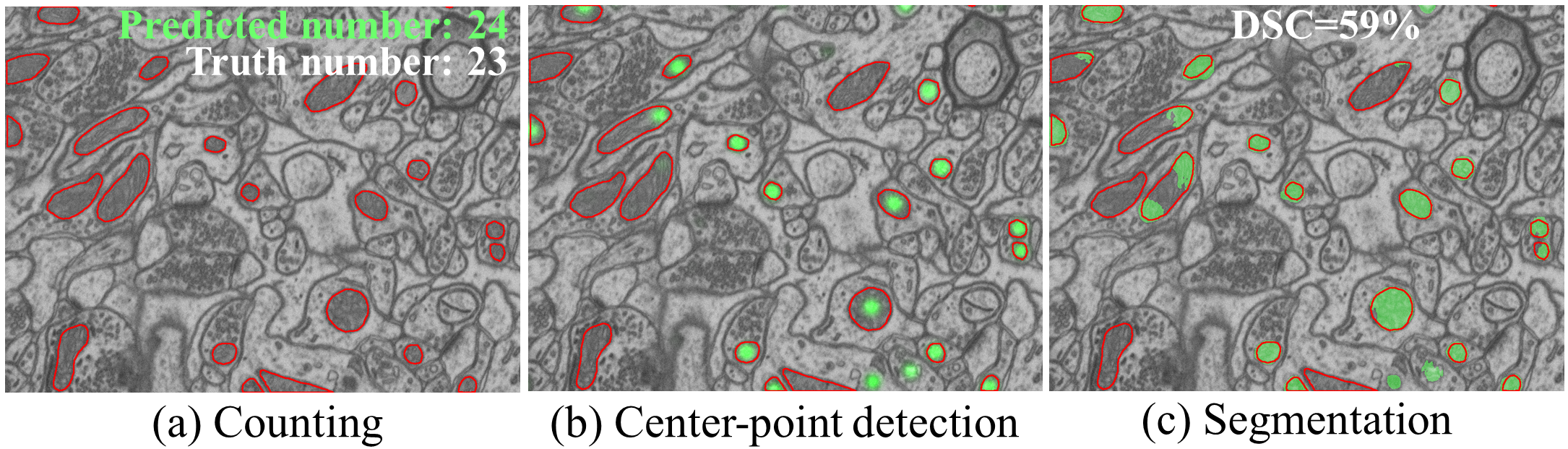}
\caption{Comparison of source models of different tasks when directly applying on the target domain. Red: ground truth; Green: model predictions. } \label{fig:2}
\end{figure}

To address the domain-adaptive segmentation of mitochondria instances with sparse point annotations, we introduce a novel multi-task  learning framework, namely WDA-Net,  which  takes  advantage of  the correlations among \textit{instance counting}, \textit{center detection}, and \textit{segmentation}, three correlated tasks with different levels of domain invariance,  for multi-level domain alignment. A basic observation is that, given the extremely sparse center point annotation,  global counting information can enhance center point detection, while center point locations can further greatly boost the segmentation process. Moreover, with the knowledge from a related source domain, it is much easier to roughly count and locate  similar objects in an unlabeled target domain than precisely delineating object boundaries. Given these observations, we use the number estimates from a counting network trained on the source domain as a soft  global prior for the  cross-domain center-detection  task, which is further employed to guide the cross-domain segmentation process with only sparse point supervision. Moreover,  the detection  and segmentation tasks are bridged by sharing  semantic features and the  estimated background. While the center detection task with sparse point supervision is presented as a weighted center regression problem, the counting network directly predicts the number of object instances.
Moreover, we compensate the annotation sparsity  with a cross-location cut-and-paste augmentation. Various validations are conducted for comparative analysis, ablation study and  influence analysis of annotation selections.

\section{Related Work}
\textbf{Supervised segmentation.}   Compared to classical machine-learning based methods \cite{lucchi2013learning,peng2019mitochondria},   deep FCNs, especially U-Net and it's variants, have shown strikingly higher performance\cite{peng2021csnet,casser2020fast,xiao2018automatic,yuan2021hive,raza2019micro}.  Given the limited computation resources in practical applications, Peng  \textit{et al.}  \cite{peng2021csnet}  introduced a lightweight 2D  CS-Net with  novel hierarchical dimension-decomposed (HDD) convolutions  and obtained state-of-the-art (SOTA) performance for both mitochondria  and nuclei segmentation.

\textbf{Weakly-supervised segmentation.}
 Full point annotation as weak labels  has been considered  \cite{Nishimura2021,2020Weakly} for cell image segmentation.  For  nuclei segmentation in histopathology images, Qu \textit{et al.} \cite{hui2020weakly} also considered partial point annotation and used an extended Gaussian mask and self-training to learn more center points and backgrounds. Then, they explored pseudo labels estimated from Voronoi partition and  clustering and used a dense conditional random field (CRF) loss for refinement.  Chen \textit{et al.} \cite{chen2021weakly} utilized  sparse points on both  nuclei  and the background and  converted the task into super-pixel classification with  a few super-pixels labeled. Compared to the  densely distributed nuclei, mitochondria are usually  sparsely distributed with large shape variance in EM images, making the Voronoi partition/clustering/super-pixel segmentation be less efficient. Different from previous studies, we focus on cross-domain segmentation with sparse point supervision.

\textbf{Domain-adaptive segmentation.}
   The main idea  behind most DA methods is to minimize the discrepancy between the feature distributions of the source  and target data by various strategies, such as adversarial learning \cite{Outputspace}  and reconstruction learning\cite{DeepG}. The  AdaptSegNet in \cite{Outputspace} conducted domain alignment in the label space with adversarial learning and has shown SOTA  performance in many tasks. For cross-domain EM image segmentation, the Y-Net  in \cite{Y-Net}  learned
  domain invariant features that can reconstruct  both the source  and target images.  Peng  \textit{et al.}  \cite{peng2020unsupervised} introduced DAMT-Net and  has obtained SOTA performance by exploring  multi-space domain alignment.  Another class of strategy is to conduct self-training on pseudo labels of the domain data \cite{2018Domain}, which does not directly conduct domain alignment. While self-training based methods show promising results, they highly rely on the domain gap and the strategy to select confident pseudo labels.

For cross-domain multi-class segmentation, Paul  \textit{et al.}  \cite{Paul2020ECCV} considered image-level labels and per-class point annotation. However, category information  is not informative for  binary mitochondria segmentation.
Xu  \textit{et al.}  \cite{2021Box} investigated bounding-box annotations  and  obtained  impressive results for cross-domain liver segmentation. However, it is still laborious to manually annotate the bounding-boxes of  a large amount of  object instances in EM images.  In contrast, the sparse point annotation is much easier to collect, even by non-experts. To the best of my knowledge, we are the first  study that uses sparse point annotation for domain-adaptive segmentation.

\section{Method}
To achieve accurate cross-domain segmentation with minimal annotation cost, we investigate the domain adaptation with extremely sparse center-point annotations.
In the considered WDA setting, we have an almost unlabeled target domain $\mathcal {D}_t$ with only sparse point annotation  $\bar{c}^t \in \{0,1\}^{H,W}$, taking 1 only at the center of a few  mitochondria in the target image  $x^{t}\in \mathbb{R}^{H,W}$ as shown in Fig. \ref{fig:1} (d). $H$ and $W$ are the image height and width, respectively. Moreover, we are provided with a fully labeled source domain $\mathcal{D}_s$ with per-pixel label $y^s\in \{0,1\}^{H,W}$ for each source image $x^{s}$. Additionally, we also define an auxiliary source label image  $c^s\in \{0,1\}^{H,W}$, which takes 1 at the mass center of each mitochondrion in  $x^{s}$.

\begin{figure}[t]
\centering
\includegraphics[width=0.49\textwidth]{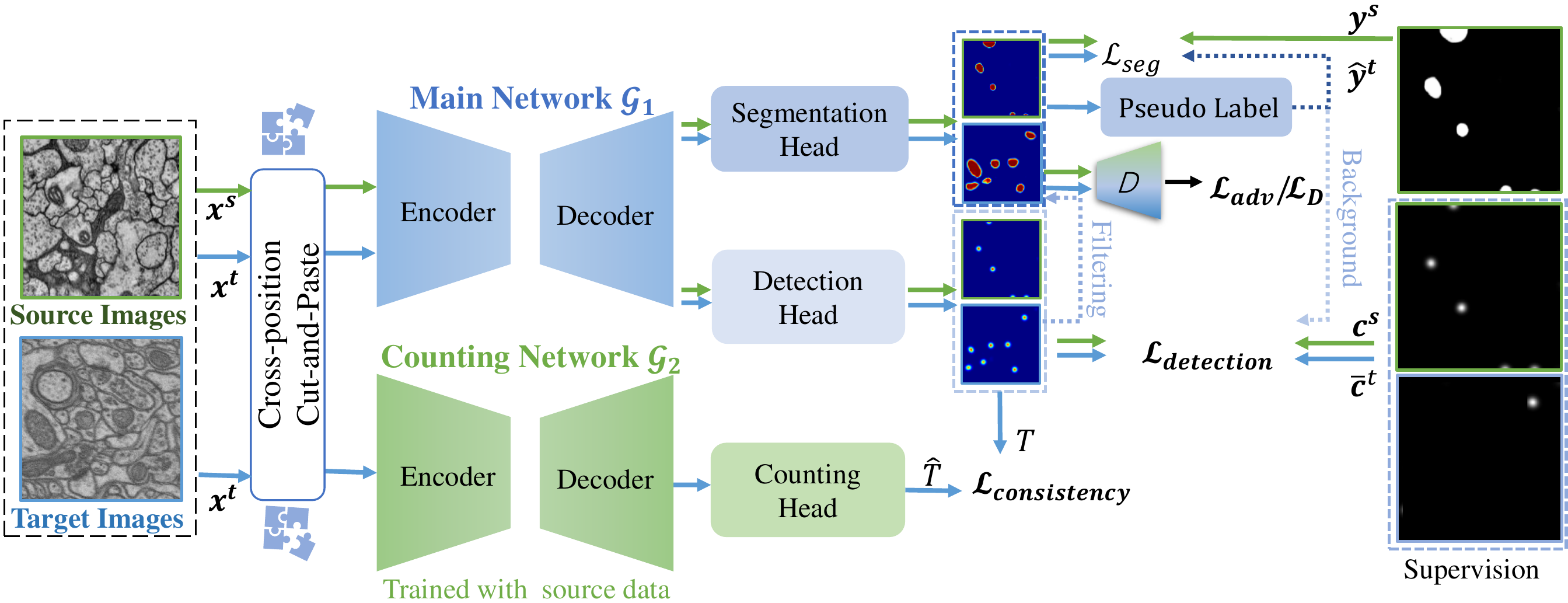}
\caption{Overview of the proposed WDA-Net.
 Three complementary tasks with different levels of domain invariance are conducted. The auxiliary counting task guides  the center-point detection task, which helps  locate mitochondria and filter out false positives.
  } \label{fig:3}
\end{figure}

Figure \ref{fig:3} presents an overview of  the proposed WDA-Net, which is a multi-task deep network that takes advantage of the relations among counting, center detection, and segmentation, three closely-related tasks but with different levels of domain invariance.
The WDA-Net comprises a detection-segmentation network $\mathcal{G}_1$  and an auxiliary counting network $\mathcal{G}_2$.  The network $\mathcal{G}_1$ is a two-head convolutional network, consisting of a segmentation head that predicts the segmentation map $p$ for each input image $x$, and a center detection head that predicts a heat-map $\hat{d}$, peaking
at centers of  mitochondria in the input image  $x$.  Since the counting head lacks full supervision signal under the sparse point annotation, we introduce a rough counting prior through the counting network $\mathcal{G}_2$, which directly predicts the number of mitochondria and is pre-trained only on the source domain. Despite the existence of domain shift, the prediction of the counting network with sufficient data augmentation can still provide useful guidance to the heatmap regression in the detection task, especially at the early training stage. Intuitively, roughly counting the number of the object instances with source domain knowledge is much easier than pixel-wise  all objects. An example is shown in Fig. \ref{fig:2}.   Therefore, we enforce a soft consistency constraint between the detection and counting predictions.

 \subsection{The Segmentation Task}
The segmentation head of the WDA-Net takes advantage of both pseudo-label learning and adversarial learning to enhance the supervision from the target domain and  minimize the domain discrepancy. Given a source segmentation model, we explore pixel-wise labels in the source domain and  pseudo-labels in the target domain to capture more complete content of the mitochondria. The corresponding loss is written as,
\begin{equation}\label{eq:1}
\mathcal{L}_{s}=\frac{1}{|\mathcal{D}_s|}\sum_{x^s} L_{ce}(p(x^s),\mathbf{y}^s)+\frac{1}{|\mathcal{D}_t|}\sum_{x^t} L_{ce}(p(x^t),\hat{\mathbf{y}}^t)
\end{equation}
where $L_{ce}$ denotes the cross-entropy loss function, $p(\cdot)$ denotes the probability outputs of the segmentation head, $\mathbf{y}^s\in \{0,1\}^{H,W,L}$ ($L$=2) is the one-hot encoding of  $y^s$ and takes an one-hot vector at each pixel $x^s_i$, and $\hat{\mathbf{y}}^t$ is the one-hot representation of the pseudo-labels  for the target image $x^t$ and takes $\textbf{0}$ at regions with null pseudo-labels. While the source labels are available,   pseudo-labels are generated with self-training to approximate true labels in the target domain.

\textbf{Pseudo-label learning.}
To  boost the segmentation with the unlabeled target data, we augment the segmentation pipeline with domain-adaptive  pseudo-labeling in the target domain (the second term in Eq. \ref{eq:1}), which selects predictions with high confidence on unlabeled regions as pseudo-labels.
Rather than simply thesholding the prediction $p(x^t)$ for extracting pseudo-labels \cite{peng2021medical}, we select pseudo-labels by exploring entropy \cite{lee2013pseudo,saporta2020esl,peng2021medical} and introduce an entropy-based criteria.  The  pseudo-label $\hat{\mathbf{y}}^{t}$ at the $i$th pixel for the class $l$  is estimated as follows,
\begin{equation}
\hat{\mathbf{y}}_{i,l}^{t}=\left\{\begin{array}{lc}
1, &\mathrm{if} ~l=\arg\max_{\tilde{l}}~ p^{t}_{i,\tilde{l}}\ \mathrm{and}\ E(p^{t}_{i})<v_{l} \\
0, &\rm{otherwise}
\end{array}\right.
\end{equation}
where $p^{t}_{i,l}$ is the abbreviation of $p(x^t)_{i,l}$,  $p^{t}_{i}=[p^{t}_{i,1}, p^{t}_{i,2}, \cdots, p^{t}_{i,L}]^T$, $E(\cdot)$ denotes the entropy, and $v_l$ is a threshold over the entropy score for the $l$ th class,
\begin{equation}
v_l= D_K \{E(p^{t}_{i}) \mid x^{t}_i \in \mathcal {D}_t, l=\mathrm{\arg \max}_{\tilde{l}} ~p^{t}_{i,\tilde{l}}\}
\end{equation}
where $D_K$ denotes the $K$th decile. In our experiment, we set $K$=8 and select  the top 80\% most confident label measured in the entropy as pseudo labels.

\textbf{Adversarial learning.}  Since domain shift will lead to degraded performance when directly applying the source model on the target data, we employ adversarial learning \cite{Outputspace} in  the  segmentation output-space. Concretely, a discriminator $D$ formed as a fully convolutional network is imposed to  distinguish whether the input  is  the prediction of a source image  or  a target image. For model training, we alternatively  train a domain discriminator ($D$) by minimizing a discriminator loss $\mathcal{L}_{D}$  \cite{Outputspace}, and update $\mathcal{G}_1$ jointly with the segmentation loss $\mathcal{L}_{seg}$ and the additional guidance from an adversarial loss $\mathcal{L}_{adv}$  \cite{Outputspace}, which  minimizes the  distribution discrepancy between the target domain and the source domain.
The adversarial learning helps adapt the source model to the target data through learning domain-invariant features and also provides a relatively sound model for initializing pseudo-labeling.

\subsection{Center Detection with Sparse Point Supervision}
To take advantage of the spare point supervision $\left\{\bar{c}^t\right\}$ on the target domain,  we train an auxiliary regression-based detection head, which can predict activation maps for  mitochondria locations.  The detection task influences the segmentation both in implicit and explicit ways. First, the detection network influences  the segmentation network by  sharing most feature layers. Second, the predicted activation maps help discover informative pixels for the segmentation and  the peaks of the activation maps indicate the locations of mitochondria instances.   The  loss function for detection is defined as follows,
\begin{equation}\label{Eq:6}
\begin{aligned}
\mathcal{L}_{d}=&\frac{1}{|\mathcal{D}_s|}\sum_{x^s,i}\left(1+\lambda \beta_{i}^{s}\right)\left(\hat{d}_{i}^{s}-d_{i}^{s}\right)^{2}+\\
&\frac{1}{|\mathcal{D}_t|}\sum_{x^t,i} \left(w_{i}+\lambda \beta_{i}^{t}\right)\left(\hat{d}_{i}^{t}-\bar{d}_{i}^{t}\right)^{2}
\end{aligned}
\end{equation}
where $d^{s}$ is the ground truth heatmap ($d^{s}=G_{\sigma_1}*c^s$) of $c^s$,  $G_{\sigma_1}$ is the Gaussian kernel with  bandwidth $\sigma_1$,  and  $\hat{d}^{s}$ denotes the estimated heat map for  $x^s$; $\bar{d}^{t}$ is the ground truth  heatmap ($\bar{d}^{t}=G_{\sigma_1}*\bar{c}^t$) of the sparse center-annotations $\bar{c}^t$,  and  $\hat{d}^{t}$ denotes the estimated heatmap for the target image $x^t$.  Since we only have labels on partial center points in the target domain, we include a spatial weight map $w$,  only taking positive values on the estimated background (i.e., the regions with $p^{x^t}_{i,1}< \rho$, where $\rho$ is set as 0.1 in the experiments) from the segmentation prediction and the estimated foreground (i.e., the regions with $\bar{d}_i^{t}> 0$). For the pixel $i$,
\begin{equation}
w_{i}=\left\{\begin{array}{cc}
1 & ~~p^{x^t}_{i,1}< \rho ~ \mathrm{or}~ \bar{d}_i^{t}> 0 \\
0 & \rm otherwise
\end{array}\right.
\end{equation}
Thus, for a target image, the loss in Eq. (\ref{Eq:6}) is essentially computed on partial regions, neglecting regions with high uncertainty.
Since we have high confidence about the true label in a small neighborhood of the labeled center points,  we include an additional weight $\lambda \beta$ in Eq. (\ref{Eq:6}) with $\beta^s=G_{\sigma_2}*c^s$ and  $\beta^t=G_{\sigma_2}*\bar{c}^t$. In our experiment, $\sigma_2$ is set to be smaller than $\sigma_1$ since we have high confidence in a small neighborhood of each annotated point. The detection prediction is also used to filter out false positives with connected component analysis.

\subsection{Counting as a Global Prior}
While the detection head is designed to  locate mitochondria centers in the target domain, the cross-domain training procedure  lacks  constraint with sparse point annotation  only for the foreground class, especially at the early training stage. To address this issue, we introduce a counting task, which involves learning global-level knowledge and  is expected to be more robust to domain shift, and regularize  the detection  with a novel counting consistency constraint.  Specifically, we utilize the labeled source data to train a counting network $\mathcal{G}_2$, which is an encoder-decoder network and initialized with model parameters from the segmentation network. Rather than regressing the  heatmap of mitochondria  locations as the detection network,  the counting network directly estimates the number of mitochondria in each input image. To improve the generalization ability of the counting model,  we use multi-scale input and diverse data augmentation.   When applying the counting model $\mathcal{G}_2$ to a target image $x^t$, we can obtain an estimated  number of mitochondria instances in $x^t$ and denote it as $T(x^t)$, which can act as a rough global prior.  Given the predicted activation map $\hat{d}^{t}$ of the detection branch of the main network $\mathcal{G}_1$ for $x^t$, we can also  obtain an estimated  number of mitochondria instances in $x^t$ and denote it as $\hat{T}(x^t)$.  In the ideal case, $\hat{T}(x^t)$ will equal to $T(x^t)$. However, there may be discrepancy between $T(x^t)$ and the ground truth due to inaccurate estimation by $\mathcal{G}_2$. Thus, we introduce a consistency loss with a small soft margin $\varepsilon$ (3 in our experiment) to enforce soft consistency between counting and detection predictions.
\begin{equation}
\mathcal{L}_{c}=\frac{1}{|\mathcal{D}_t|}\sum_{x^t} \max \left(0,(T-\varepsilon)-\hat{T}\right)+\max \left(0, \hat{T}-(T+\varepsilon)\right)
\end{equation}

\subsection{Cross-Position Cut-and-Paste Label Augmentation }
A significant challenge of  exploring supervision in the target domain is the extremely sparsity of the point annotations. To compensate for this, we introduce a cross-position cut-and-paste augmentation (CP-Aug) strategy similar to cutmix \cite{yun2019cutmix}. However, the cropped  patches are not necessarily pasted in the same position with the aim to  increase the density of point annotations.  Given two images $(x^t_A, \bar{c}^t_A)$ and $(x^t_B, \bar{c}^t_B)$ from the target domain, we aim to generate a new image $(x^t_C, \bar{c}^t_C)$ by cutting a rectangular region (256$\times$256 in our experiments) with largest number of annotation points in $x^t_A$ and pasting it to a rectangular region  of the same size but with few number of annotation points  in $x^t_B$. The synthesized images have more annotated points and  will  be used in model training.
\subsection{Overall Optimization and End-to-End Learning}
For model training, we alternatively  train the domain discriminator ($D$) by minimizing the discriminator loss $\mathcal{L}_{D}$  \cite{Outputspace}, and update $\mathcal{G}_1$.
Specifically, given the discriminator $D$ and pretrained counting network $\mathcal{G}_2$, we train the main network $\mathcal{G}_1$ by minimizing the following loss,
\begin{equation}
\mathcal{L}_{obj}=\mathcal{L}_{s}+\lambda_a\mathcal{L}_{adv}+\lambda_d\mathcal{L}_{detection}+\lambda_c\mathcal{L}_{c}
\end{equation}
where $\lambda_a$, $\lambda_d$ are trade-off parameters; $\lambda_c$=1-$z$/$z_{max}$ decays along with  iteration $z$, and $z_{max}$  is the maximum iteration.

\section{Experiments}
\subsection{Dataset and Evaluation Settings}

\textit{Drosophila  Data} \cite{DrosophilaIIIVNC}  were taken from Drosophila melanogaster third instar larva VNC using serial section Transmission Electron Microscope (ssTEM) in  an anisotropic resolution of 4.6$\times$4.6 $nm$/pixel with the section thickness of 45-50 $nm$.
This image stack is of size 20$\times$1,024$\times$1,024.

\textit{EPFL  Data} \cite{lucchi2013learning} were taken from the hippocampus of a mouse  using  focused ion beam scanning electron microscope (FIB-SEM) in  a resolution of 5$\times$5$\times$5 $nm$.  Both the training and testing subsets are of size  of 165$\times$768$\times$1,024.

We evaluate our method on the adaptation from the small Drosophila dataset to the large EPFL  dataset.

\textbf{Evaluation metrics.} We utilize both class-level measure, i.e., Dice similarity coefficient (DSC) in \%, and instance-level measures, i.e., Aggregated Jaccard-index (AJI) \cite{kumar2017dataset}  in \% and Panoptic Quality (PQ) \cite{kirillov2019panoptic}  in \%.

\textbf{Network architectures.} For the backbone of the  $\mathcal{G}_1$ network, we use a lightweight variant of U-Net, which uses the HDD unit  in \cite{peng2021csnet} as the basic building blocks.
Besides the prediction layer, the segmentation head and detection head use one HDD layer and two HDD layers, respectively. To obtain  counting  prediction from  the detection output, we utilize one HDD  layer  followed by an integration layer for prediction.  The counting network $\mathcal{G}_2$  has the same backbone as $\mathcal{G}_1$ with a integration layer for final prediction. We use the same discriminator as \cite{peng2020unsupervised}, which is a fully-convolutional network of 5 layers with   channel
numbers of \{64, 128, 256, 512, 1\}.

\textbf{Parameters and training settings.} We set $\lambda_a$=$10^{-3}$, $\lambda_d$=$10^{-1}$, $\lambda$=$3$, and $\beta=0.2$. The bandwidths $\sigma_1$ and $\sigma_2$ are set as 10 and 2, respectively.
We implement the models  on one 1080Ti GPU. The $\mathcal{G}_1$ network is trained using SGD with initial learning rate $5\times 10^{-5}$ and polynomial decay of power 0.9. The  $z_{max}$ is set to 10k and the batch size is 1. We randomly crop patches of size 512$\times$512 as training input. Data augmentations including flipping/rotation, blur, color jitter, and the proposed CP-Aug are used. The discriminator is trained with  Adam optimizer as \cite{peng2020unsupervised}. The $\mathcal{G}_2$ network is initialized with parameters of the $\mathcal{G}_1$ network and optimized with mean squared loss and  Adam optimizer. Multi-scale inputs of size 512 $\times$ 512, 768 $\times$ 768, and 1024 $\times$ 1024 obtained through crop and resampling  and data augmentation including  flipping/rotation, blur, and color jitter are used for training the  $\mathcal{G}_2$ network.

\begin{table}[t]
\caption{Ablation study of  the proposed WDA-Net (15\%).}
\centering
\label{tab:1}
\setlength{\tabcolsep}{1.5mm}
\begin{tabular}{lccccccc}                  \\
\hline
&&&&&&\multicolumn{2}{c}{Drosophila  $\rightarrow$EPFL}\\
 \cline{7-8}
Model     &Detect. & Count& P-L & CP-Aug&Filter & DSC(\%)  & PQ(\%)\\\hline
I            & ~        & ~           & ~ & &       & 71.0& 45.6  \\
II       & \checkmark        & ~    & ~& ~   &        & 84.6& 60.8    \\
III   & \checkmark        & \checkmark    & ~  & ~  & & 85.9& 65.0 \\
IV         & ~        & ~    & \checkmark & ~   &     &77.5  & 48.4  \\
V      & \checkmark        & ~    & \checkmark  & ~  && 87.6 & 64.7    \\
VI & \checkmark& \checkmark    & \checkmark  & ~   &   & 89.3 & 70.4  \\
VII  & \checkmark   & \checkmark    & \checkmark  & \checkmark & & 90.5 & 73.3    \\
%VII  & \checkmark  & \checkmark & \checkmark& \checkmark  &\checkmark& &92.1&   73.6\\
VIII  & \checkmark        & \checkmark    & \checkmark   & \checkmark  &\checkmark& \textbf{90.7}         & \textbf{76.5}\\
\hline
\end{tabular}
\end{table}

\subsection{Ablation study}
 Table \ref{tab:1} demonstrates   the individual  contributions of our key  components, a) Detect.: using the detection head; b) Count: using the counting consistency constraint to the detection head; c) P-L: seudo-label learning; d) CP-Aug:  the cross-position cut-and-paste augmentation, e) Filter: connected component analysis of the segmentation with points from the detection head  and also removing noise blobs with open-and-close operation. The ablation studies are conducted  using 15\% point annotation, namely WDA-Net (15\%), for Drosophila $\rightarrow$EPFL.

While the baseline Model I only conducts UDA with adversarial training \cite{Outputspace}, sequentially adding the key components  results in  gradually improved performance. By integrating the center detection to the Model I, we achieve  a  significant improvement of 13.6\% in DSC, 15.2\% in PQ. By introducing the counting prior  constraint, we obtain further improvement, especially a large  performance gain (4.2\% in PQ)  in detection. By comparing Model V, Model IV, and Model I, we observe that integrating the detection with  the pseudo-label learning can significantly improve both the segmentation  and  detection tasks, while the performance gain in PQ by only using  pseudo-label learning is limited.  By comparing Model VII and Model VI, we find that the CP-Aug can  lead to an improvement of  2.8\% in PQ. Finally, the detected center points can help significantly reduce false positives,  and  Model VIII outperforms Model VIII by 3.0\% in PQ. In summary, all the proposed components can consistently improve the  performance, and Detect., Count, CP-Aug, and Filter have  stronger ability to improve the detection performance.

\begin{table}[tb]
\caption{Robustness to  different selections of  15\%  sparse annotations. }
\centering
\setlength{\tabcolsep}{1mm}
\label{tab:2}
\begin{tabular}{lcccccc}
\hline
\multicolumn{7}{c}{Drosophila $\rightarrow$EPFL}                 \\
\hline
& Sample 1  &Sample 2 & Sample 3 &Sample 4&Sample 5 &Mean$\pm$Std   \\
\hline
DSC     & 90.7& 90.9 &90.4& 91.0& 91.1 &90.8$\pm$0.3    \\
PQ    & 76.5& 76.0 &75.7& 76.2 &78.9 &76.7$\pm$0.3\\
\hline
\end{tabular}
\end{table}
\textbf{Robustness to random annotation selections.} Table \ref{tab:2} shows the performance with different random selection of 15\% sparse annotations. The proposed method demonstrates robust
performance with different annotation locations, which is a desirable property for practical usage.
\subsection{Comparative Experiment}
Table \ref{tab:3}  compares WDA-Net with SOTA UDA methods, i.e., Y-Net \cite{Y-Net}, AdaptSegNet \cite{Outputspace}, and DAMT-Net \cite{peng2020unsupervised}, and the upper bound, i.e., the supervised model on the target domain. In Table \ref{tab:3}, we also compares  the proposed WDA-Net using different ratios of point annotations. Moreover, both the vanilla U-Net \cite{ronneberger2015u} and U-Net (HDD), i.e., the proposed lightweight variant of U-Net, are tested as the backbone network.

\begin{table}[t]
\caption{Comparison results on Drosophila $\rightarrow$EPFL.}
\centering
 \setlength{\tabcolsep}{1mm}
 \label{tab:3}
\begin{tabular}{lllccc}
%\hline
% \multicolumn{7}{c}{Drosophila III VNC$\rightarrow$Mouse Hippocampus}                                        \\
\hline
&&&\multicolumn{3}{c}{Drosophila $\rightarrow$EPFL}\\
\cline{4-6}
Type& Method                & Backbone & DSC & AJI   & PQ  \\
\hline
 &NoAdapt                 &  U-Net   & 57.3        & 39.6       & 26.0  \\
 \hline
 \multirow{3}{*}{UDA}&Y-Net \cite{Y-Net}  &  \multirow{3}{*}{U-Net}    & 68.2        & --       & --     \\
&AdaptSegNet \cite{Outputspace}             &    & 69.9        & --       & --      \\
&DAMT-Net \cite{peng2020unsupervised}                &     & 74.7        & --        & --    \\
\hline
%&NoAdapt                 & CS-Net   & 70.1     & 54.4     & 53.5     & 46.4     \\
 \multirow{3}{*}{UDA}&Y-Net \cite{Y-Net}                   & \multirow{3}{*}{U-Net (HDD)}    & 69.6        & 52.2     & 42.6   \\
&AdaptSegNet \cite{Outputspace}             &   & 71.2       & 54.9     & 47.3   \\
&DAMT-Net \cite{peng2020unsupervised}               &     & 75.3         & 59.7     & 47.7  \\
\hline
 \multirow{4}{*}{WDA}&Our model (5\%)      &  \multirow{6}{*}{U-Net (HDD)}
                            & 88.5& 79.3 &74.5     \\
&Our model (15\%)      &    & 90.7& 82.1& 76.5      \\%%
&Our model (50\%)       &   & 91.0& 83.4 &77.8    \\
&Our model (100\%)      &  & 91.6& 83.9 &78.3    \\
\hline
Upper Bound &       Supervised      &  U-Net (HDD)  & 93.6       & 87.9     & 80.2  \\
\hline
\end{tabular}
\end{table}

First, we can observe that the U-Net with HDD \cite{peng2021csnet} shows better performance than the vanilla U-Net, while the model size  (6.7M) of the U-Net (HDD) is only about 1/5 of the model size (34.5M) of the vanilla U-Net. Thus, in the following experiment, we use the U-Net (HDD) as the backbone. Second,  the well-trained source models show significant performance drop when directly applying on the target domain, which indicates the severe domain shift and the sensitivity of deep network models.

Third, with extremely low annotation cost, our WDA-Net  significantly outperforms the UDA counterparts and shows comparable  performance with the supervised trained counterpart. By comparing the performance of our model with annotation ratios, we find that annotating on  15\%  instances of the target training data can already produce sound performance on the testing data. The proposed WDA-Net (15\%) outperforms the DAMT-Net by a large margin with only minimal annotation cost and obtains a performance of 92.8\% in DSC and 78.3\% in PQ, which are only 2.0\% and 1.9\% lower, respectively,  than that of the supervised model.

\begin{figure}[t]
\centering
\includegraphics[width=0.5\textwidth]{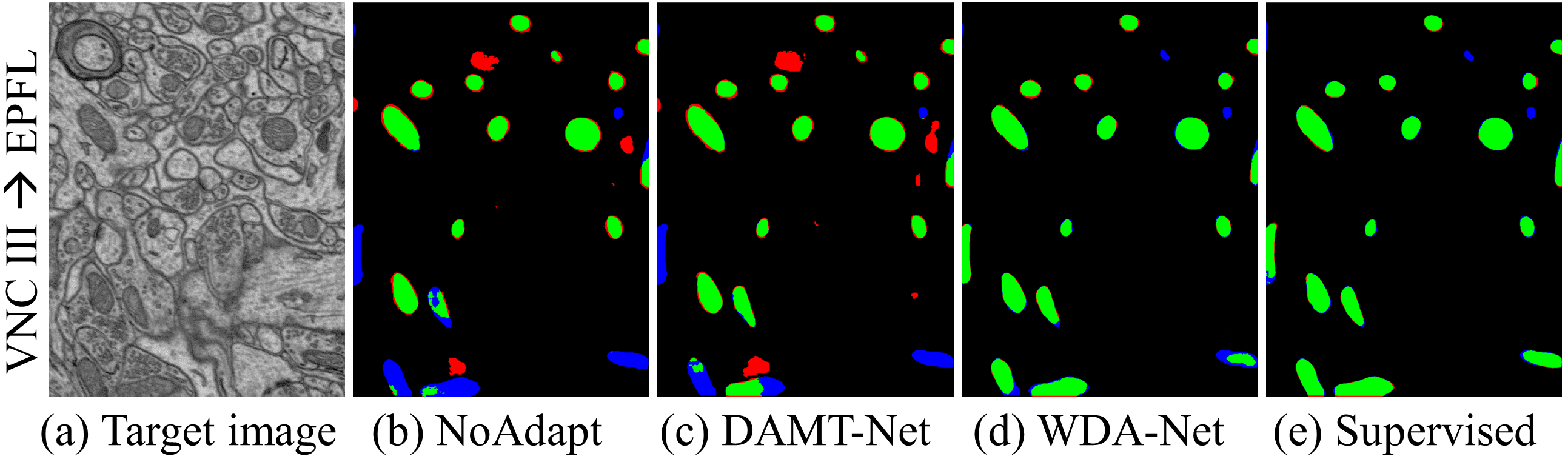}
\caption{Visual comparison of the proposed WDA-Net (15\%) with other methods. Red:  false positives;  Blue: false negatives; Green:  true positives.  } \label{fig:4}
\end{figure}
Figure \ref{fig:4}  visually compares our WDA-Net with other methods. Compared to NoAdapt, the DAMT-Net can recognize more mitochondria instances similar to the source  mitochondria, but this method still has many false positives and false negatives. In contrast, with minimal annotation effort, the proposed method can significantly reduce false positives and false negatives, and shows comparable results with the fully-supervised method. Moreover, with 15\% center points for the training data, our method can already produce sound results.

\section{Conclusion}
In this study, we addressed domain-adaptive mitochondria segmentation  under sparse center point supervisions. We introduced a task pyramid learning framework, which introduces counting as global constraint for center detection, which further provides location information for the segmentation. A novel cross-position  cut-and-paste strategy is introduced to compensate for the sparsity of the partial point annotation.  Experiments on challenging benchmarks have showed that our model can produce sound performance close to the supervised counterpart with only 15\% partial point supervision. The validations also showed the robustness of our model.  In future work, we will consider  source-free setting and explore only  well-trained source models and the target data for adaptation.


\end{document}